\documentclass[letterpaper, 10 pt, conference]{ieeeconf}
\IEEEoverridecommandlockouts
\usepackage{amsmath,amsfonts}
\usepackage{algorithmic}
\usepackage{algorithm}
\usepackage{array}
\usepackage[caption=false,font=normalsize,labelfont=sf,textfont=sf]{subfig}
\usepackage{textcomp}
\usepackage{stfloats}
\usepackage{url}
\usepackage{verbatim}
\usepackage{graphicx}
\usepackage{cite}
\usepackage{array}
\usepackage{multirow}
\usepackage{xcolor}
\usepackage{gensymb}

\usepackage{amssymb} 
\usepackage{booktabs}
\begin{document}

\title{ARCSnake V2: Mechanical Adaptations For An Amphibious Multi-Domain Screw-Propelled Snake-Like Robot}

\author{Sara Wickenhiser$^{1}$$^{\dagger}$, Lizzie Peiros$^2$$^{\dagger}$, Calvin Joyce$^{1}$, Peter Gavrilov$^1$, Sujaan Mukherjee$^1$, Syler Sylvester$^2$,\\ Junrong Zhou$^1$,  Mandy Cheung$^2$,  Jason Lim$^2$, Florian Richter$^2$, Michael C. Yip$^2$ \IEEEmembership{Senior Member,~IEEE}%
\thanks{$^{\dagger}$ Equal contribution.}
\thanks{$^1$Mechanical and Aerospace Engineering Department, University of California San Diego, La Jolla, CA 92093 USA {\tt\footnotesize\{swickenhiser, cajoyce, pgavrilo, 
s2mukherjee, 
juz058\}@ucsd.edu}}%
\thanks{$^2$Electrical and Computer Engineering Department, University of California, San Diego, La Jolla, CA 92093 USA. {\tt\footnotesize\{epeiros, m3cheung, ssylvester, jkl009, frichter, m1yip\}@ucsd.edu}}
}


\maketitle


\begin{abstract}

Robotic exploration in extreme environments, such as caves, oceans, and extraplanetary surfaces, poses significant challenges, specifically locomotion across diverse terrains. Conventional wheeled or legged robots often struggle in such contexts due to variability in traction. This paper presents ARCSnake V2, an adaptation of ARCSnake V1 with additional amphibious capabilities for aquatic environments. ARCSnake V2 combines the high mobility of hyper-redundant snake robots with the terrain versatility of Archimedean screw propulsion. Key contributions include a water-sealed mechanical design with serially linked screw and joint actuation, and an integrated buoyancy control system. Extensive experiments validate its underwater capabilities for diving and surfacing, as well as force-regulated actuation. These capabilities position ARCSnake V2 as a versatile platform for exploration, search-and-rescue, and environmental monitoring in multi-domain settings.

\end{abstract}




\section{Introduction}

Robotic exploration in extreme environments, such as caves, oceans, and ice found on extra-planetary surfaces, poses significant challenges \cite{kalita2020exploration}.
Wheeled robots, while energy-efficient on flat surfaces, struggle with uneven or soft terrains due to limited traction and obstacle negotiation capabilities \cite{bruzzone2012locomotion}.
Legged robots alike, quadrupeds, humanoids, etc. face difficulties in navigating and maneuvering in unpredictable terrain, particularly those that even humans have not mastered without specialty tools, which limit their practical deployment in such environments \cite{motionplanning2024}. Even more restrictive are environments which can only be reached by robots capable of traversing tight spaces. Applications such as search and rescue, environmental monitoring, and space exploration demand robots that can adapt to varying terrains and overcome obstacles that traditional form factors cannot handle effectively \cite{li2022alternative, pena2020collaborative}.

\begin{figure} [t!]
    \centering
    \includegraphics[trim={0 1.75cm 0 0},clip, width=\linewidth]{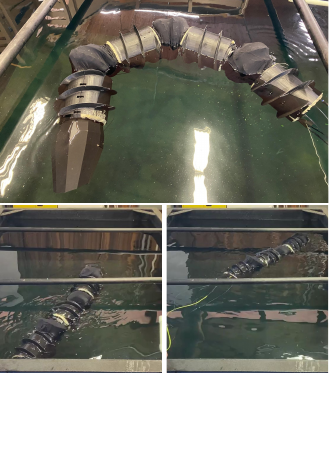}
    \caption{ARCSnake is a water-sealed, amphibious, screw-propelled, snake-inspired robot for traversing multi-domain environments. It is capable of aquatic motion as seen by the starting and ending points of a swim in the bottom left and right images respectively. It is equipped with a buoyancy control system for underwater stability. }
    \label{fig:cover_figure}
\end{figure}

A promising approach to address these challenges is the development of multi-domain mobile robots that combine different locomotion strategies. One such design is ARCSnake\cite{arcsnake_icra}\cite{arcsnake_tro}, the precursor to EELS\cite{carpenter2021exobiology}, a screw-propelled, snake-like robot that integrates the advantages of screw propulsion with a hyper-redundant, snake-like backbone. Rotating Archimedean screws have proven effective for propulsion in amphibious and granular media \cite{neumeyer1965marsh, dugoff_1967,chen2025characterization} while snake-like robots offer unique locomotive abilities and a compact form factor ideal for navigating confined spaces similar to vine robots \cite{coad2019vine}.

In this work, we present major mechanical upgrades of our screw-propelled snake-like robot, ARCSnake V2, which provide water-proof capabilities. 
ARCSnake V1 demonstrated many impressive capabilities, but its major limitation was its lack of underwater capabilities.
While the core concept of using repeated segments paired with U-Joints and screw modules remains unchanged, we present several key design improvements centered around water-based locomotion.
The main technical contributions are as follows: a waterproofed mechanical design for a serially linked screw drive train and cable-driven joint actuation, and a selective, adaptive buoyancy system for depth control and underwater maneuverability.
Each segment is rigorously tested, providing the equivalent to IP67 rating for our proposed system.

\section{Related Works}





 
Snakes exhibit a remarkable ability to traverse complex and unstructured terrains despite their structurally simple bodies, which has inspired extensive research into snake-inspired robotic systems that replicate their morphology and locomotion strategies \cite{hirose_1992, hirose2004biologically, transeth2009survey}.
The most common snake robot designs consist of repeated modular segments connected by actuated joints, forming a hyper-redundant system \cite{wright2007design}.
Early designs employed a single degree of freedom per segment to achieve planar serpentine locomotion \cite{hirose2009tutorial}, with some systems demonstrating amphibious capabilities, including surface locomotion on water \cite{crespi2006amphibot, kamamichi2006snake}.
The natural progression toward higher-degree-of-freedom joints \cite{shammas2006three} significantly expanded functionality, enabling behaviors such as stair climbing \cite{nakajima2018motion}, pipe climbing \cite{trebuvna2016inspection}, and sidewinding on loose sand \cite{marvi2014sidewinding, gong2014learning}. These additional degrees of freedom also facilitated improved control of normal forces for sustained ground contact, a key challenge identified in early snake robot designs \cite{shugen2001analysis, ma2006analysis, bayraktaroglu2006design}.

Despite increased kinematic complexity, early snake robots failed to replicate the anisotropic friction that biological snakes generate via direction-dependent ventral scales to produce forward thrust.
This observation motivated the development of passive skin mechanisms designed to mechanically induce frictional anisotropy, including passive wheels \cite{ye2004turning,crespi2005amphibot,wu2010cpg, cao2017robust}, parallel grooves \cite{saito_fukaya_iwasaki_2002}, fins \cite{hirose2009tutorial}, and skates \cite{endo1999adaptive}.
Building upon this foundation, later work incorporated sensory feedback to improve robustness and environmental adaptability, enabling strategies such as adaptive stiffness control \cite{zhang2020environmental}, adaptive sidewinding \cite{marvi2014sidewinding}, and obstacle-aided locomotion \cite{transeth2008obstacle, liljeback2011experimental}.


While passive skins introduce directional friction, they fundamentally couple propulsion to body deformation, limiting independent control of contact and thrust.
To address this limitation, researchers developed active skin mechanisms in which actuated surface elements directly generate propulsion, effectively decoupling thrust production from body kinematics; these implementations include active wheels \cite{klaassen1999gmd,date2000locomotion}, omnidirectional wheels \cite{ye2009modular}, rotating bodies with passive rollers akin to Mecanum configurations \cite{fukushima2012modeling}, tank treads \cite{armada2005omnitread,borenstein2007omnitread}, and toroidal skins \cite{mckenna2008toroidal}.

Among active surface propulsion mechanisms, screw-based designs represent a distinct approach that generates thrust through continuous helical engagement with the environment. Early screw-propelled vehicles were developed for traversing snow, marshes, sand, and mud \cite{neumeyer1965marsh, dugoff_1967}, demonstrating the ability of rotating helical blades to generate thrust through continuous interaction with compliant media. Subsequent experimental and computational studies examined the mechanics of screw–terrain interaction in granular substrates \cite{Cole_1961, dugoff1967model, lim2023mobility}, with modeling efforts grounded in terramechanics principles \cite{Cole_1961, nagaoka2010terramechanics}. These works identified key geometric parameters, particularly lead angle and blade height, as dominant factors influencing propulsion efficiency \cite{Cole_1961, nagaoka2010terramechanics, joyce2023nasu}. The most common mechanical configuration consists of two parallel counter-rotating Archimedes screw rotors \cite{dugoff1967model, fales1972riverine, nagaoka2009development, osinski2015small}, though series and quad-rotor configurations have also been explored \cite{freeberg2010study, lugo2017conceptual, OLADUNJOYE20227}.

A seminal demonstration of screw-based propulsion in snake-like robots is the ARCSnake system, which integrates Archimedean screw locomotion with a hyper-redundant serpentine architecture to achieve multi-domain mobility across varied terrains \cite{arcsnake_icra, arcsnake_tro}. The EELS robot, developed for extraterrestrial exploration, further illustrated the viability of screw-based locomotion on icy and cryogenic terrains \cite{carpenter2021exobiology}. While screw-based propulsion has shown promise in amphibious settings and controlled water experiments \cite{arcsnake_tro, lim2023amphibious}, a fully integrated, waterproofed snake-like robotic system capable of sustained underwater operation has yet to be demonstrated. Realizing such a platform requires addressing challenges in sealing distributed actuation, maintaining drivetrain integrity under hydrostatic pressure, and achieving controllable buoyancy while preserving modular articulation. Bridging this gap is essential to extending screw-propelled snake robots from multi-terrain mobility to true multi-domain operation.

\section{Amphibious Capability Design}

As was presented in \cite{arcsnake_tro, carpenter2021exobiology}, screw-propelled snake robots are capable of traversing and, in some cases, climbing ice, granular media, solid ground, and rocky terrain. Here, we will show how modifications to these designs can further enable amphibious capabilities. The specific modifications include: an IP67-rated chamber for the electronics and powered actuation that drives the screw propulsion, and a dynamic buoyancy system that enables control over depth. Additional mechanical changes were required due to the significant weight and size differences in the segments, such as u-joint force to maintain the ability to lift itself. Key mechanical and electrical components are summarized in Table~\ref{tab:system_specs}. 

\subsection{Mechanical Design: Segment Upgrades}



\begin{table}[t]
\vspace{2mm}
\centering
\caption{System Specifications}
\label{tab:system_specs}
\begin{tabular}{m{1.5cm} m{1.0cm} m{3.5cm}}
  \toprule
  \textbf{Specification} & \textbf{Within} & \textbf{Specification Details}\\
  \midrule
  \multirow{3}{3em}{Power} & Segment & 12V-48V, 240W (max)\\ 
  & System & 12V-48V, 960W (max) \\
  \midrule
  \multirow{2}{3em}{Comms} & Segment & CAN-BUS \\ 
  & System & USB and CAN-BUS \\
  \midrule
  \multirow{3}{3em}{Actuator} & Screw & Torque: 1.0 Nm Continuous, 3.8 Nm peak \\ 
  & Half U-Joint & Torque: 2.6 Nm Continuous, 13 Nm peak \\ 
  \midrule
  \multirow{3}{3em}{Dimensions} & Segment & Max Len: 0.441m, Max Dia: 0.269m \\
  & Head Segment & Max Len: 0.252m, Max Dia: 0.190m \\ 
  & System & Max Len: 1.827m, Max Dia: 0.252m\\
  \midrule
  Buoyancy & Bladders & Torus Shape\\ 
  & Shells & Foam Filled \\
  \midrule
  \multirow{2}{3em}{Pressurization} & Internal & 6 psi\\
  & Bladders & 3–5 psig, 3 psia\\
  \midrule
  Waterproof & Rating & IP67 \\
  \bottomrule
  \label{tab:system_specs}
\end{tabular}
\end{table}

\subsubsection{Screw Actuator}
\begin{figure*}[t]
    \centering
    \includegraphics[trim={0cm 4.5cm 0 0}, clip, width=.95\linewidth]{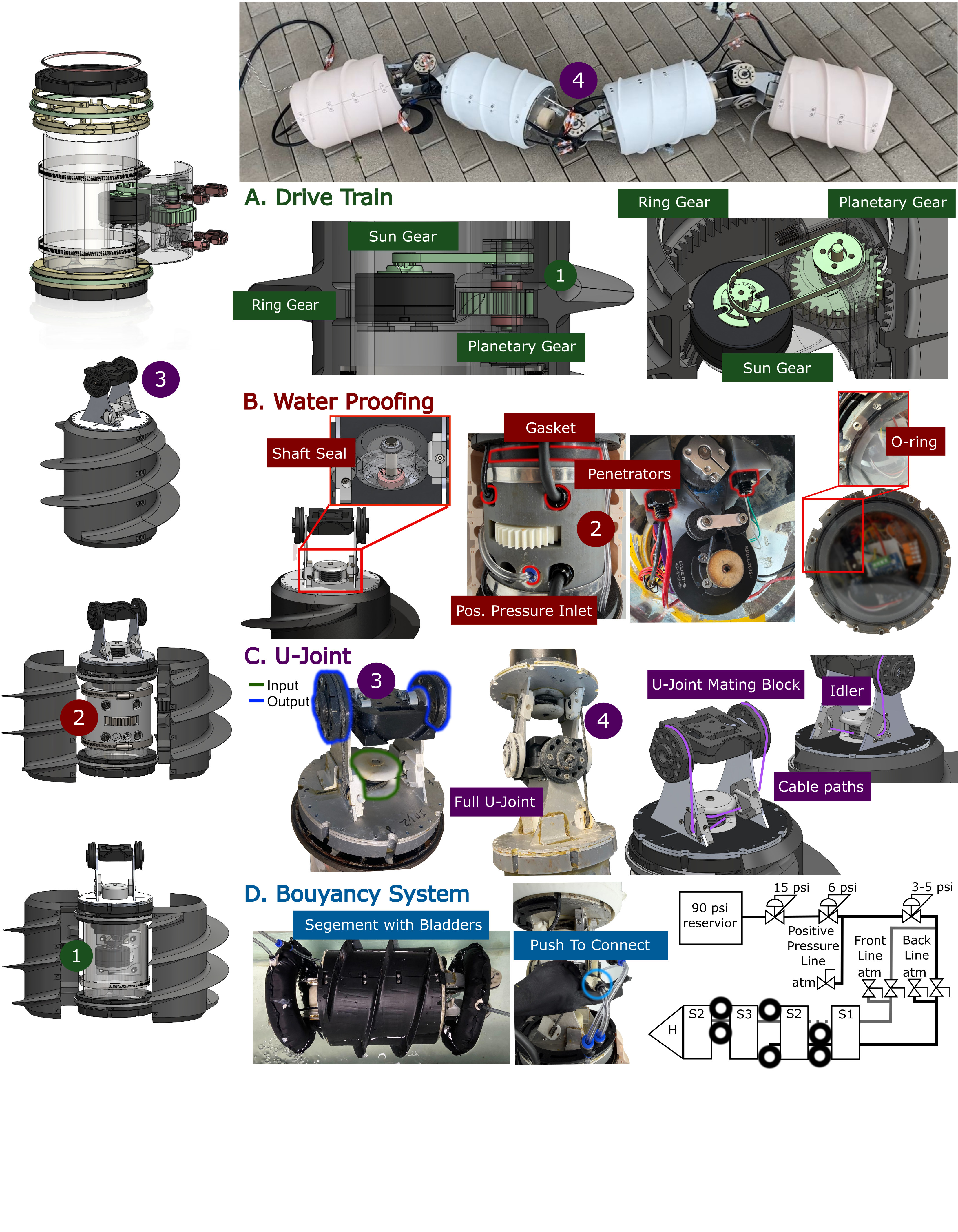}
    \caption{The top-most image shows the system fully built with its land screw shells on. The leftmost column begins with an exploded view of the major internal components of each segment, above a series of images showing the shell and the internal electronics chamber with pass-throughs. Section (A) Drive Train: show the screw drivetrain mechanism. The gear system is shown in green. The left image shows the belt drive: the sun gear mounted on the motor connected to a pulley, which in turn drives the planetary gear.  In the right image, the planetary gear meshes with the screw shell, which is the ring gear. Section (B) Water Proofing: This section, from left to right, shows the shaft seals on the planetary gear shaft, the penetrators for the power lines, the gasket seal for the screw block, the positive pressure line inlet, and the o-ring seal on each end of the inner chamber. Section (C) U-Joint: The right-most image shows a half U-joint, highlighting the motor output in green, which is the input to the pulley system at the junction where the two halves meet, depicted in the image just to the right. The final set of images, on the right, highlights the cable path over the idlers. Section (D) shows the buoyancy system with a system diagram on the right and the physical bladders seen in the images to its left.}
    \label{fig: big figure}
\end{figure*}

Starting with the transmission, from the exterior, our screw propulsion transmission begins with our 3D printed outer shell. The shell is comprised of two halves that align with each other and a set of thin 6656K17 Ultra-Thin Ball Bearings. Printed into the shells are 2 halves of a ring gear, Fig \ref{fig: big figure} section (A) in green, which connects to the planetary gear, Fig \ref{fig: big figure} section (A). This planetary gear, made of BambuLabs PLA, is driven by the sun gear, Fig \ref{fig: big figure} section (A). However, this connection must pass through the screw transmission block using Spring-Loaded Rotary Shaft Seals to seal the chamber from water, resulting in the use of an additional 5/16" diameter aluminum shaft, and an Active Robotics clamping hub (part 545624) to transfer rotation through the seal and increase the gear ratio.



Connected to the planetary gear shaft is a 3D-printed pulley with the same embedded 545624 ServoCity shaft clamp. That pulley connects to the sun gear via a 3D-printed belt-drive pulley, which is directly driven by the screw motor, an RMD-L-7015 motor.


\subsubsection{Electronics Water Proofing}

In addition to the transmission, the screw block, seen in gray in Fig. \ref{fig: big figure} section (B) in red, also serves as a conduit for the power and CAN cables. The cables from the parent screw sub-unit pass through the screw block, connect to the interior electronics, and exit the screw block at the opposite end to extend to the next sub-unit. The cable penetrators are Blue Robotics WetLink Penetrators, size 4.5 for the CAN line and 8.5 for the power line. The screw block is sealed to the internal chamber using a rubber gasket and a hose clamp (part number 45945K36).

\subsubsection{U-Joint}


To enable snake-like undulatory locomotion, angle the vector of the screw propulsion, and traverse changes in terrain elevation, ARCSnake incorporates an active universal (U-) joint at each interlink connection, see in Fig \ref{fig: big figure} section (C) in purple. The U-joint provides a hemispherical range of motion, offering over $180^\circ$ of rotation in both the pitch and yaw axes. 
In V2 each axis is actuated by a cable-driven mechanism powered by an RMD-X8 Pro motor with an internal 5:1 gearbox and an external 1.8:1 reduction, providing continuous torque sufficient to lift an adjacent link. As seen in Fig \ref{fig: big figure} section (C), the full U-joint is built from two half U-joints, which are connected through a block that aligns the two halves together with a $90\degree$ phase difference, creating the full universal joint.
Motor encoders are used for high-resolution position sensing and closed-loop control. The motor is connected to a resin-printed pulley, which the D12 M-RIG MAX coils around, allowing the motor to rotate the U-joint.

\subsubsection{Positive Pressure Line}
A positive-pressure line is fed through all segments and supplies 4 psig.
Pressurization makes all seals more robust underwater by releasing air through any cracks.
The positive pressure line uses a push-to-connect connector with an O-ring seal and an MS6M-M5 pneumatic fitting as the entrance point into the segment through the screw block, as seen in Fig. \ref{fig: big figure} section (B).

\subsubsection{Head Design}

A custom head was designed to enable sample collection and provide front-facing video data.
It is made out of BambuLabs PLA and has a custom internal 4-bar linkage mechanism that opens the mouth and simultaneously extends a gripper, powered by a 70kg IP68 servo motor.
The gripper on the ROSMASTER M3 Pro Robot is powered by a 55kg IP68 servo motor and can collect underwater samples of interest, as shown in Fig. \ref{fig:snake_head}. At the front of the head, a DJI Action 2 camera with a 135$^\circ$ field of view is positioned to capture images when the claw has accurately grasped a sample.

\begin{table}[t]
\vspace{2mm}
\centering
\caption{Mass and Volume Data}
\label{tab:MassandVol}
\begin{tabular}{m{1.7cm} m{1.25cm} m{1.4cm} m{1.3cm} m{0.95cm}}
  \toprule
  \textbf{Part} & \textbf{Mass} & \textbf{Volume} & \textbf{Density} & \textbf{Buoyancy} \\
  \midrule
  Water & 1 kg & 1 $m^3$ & 1000 $\frac{kg}{m^3}$ & 0 N\\
  \midrule
  Middle Segment & 5.386 kg & 0.00339 $m^3$ & 1583.4 $\frac{kg}{m^3}$ & -19.4 N\\
  \midrule
  Front Segment + Head & 5.006 kg + (1.009) kg & 0.004 $m^3$ & 1251.5 $\frac{kg}{m^3}$ & -39.2 N\\
  \midrule
  Tail Segment & 4.486 + (0.796) kg & 0.00318 $m^3$ & 1661.1 $\frac{kg}{m^3}$ & -20.6 N\\
  \midrule
  Marine (foam) Shells & 1.164 kg & 0.00208 $m^3$ & 559.6 $\frac{kg}{m^3}$ & 9.0 N\\
  \midrule
  Torus-Shaped Bladder & 0.02 kg & 0.00143 $m^3$ & 14.2 $\frac{kg}{m^3}$ & 13.8 N\\
  \midrule
  Regular Shells & 0.544 kg & 0.00092 $m^3$ & 591.3 $\frac{kg}{m^3}$ & 3.7 N\\
  \bottomrule
  \label{tab:MassandVol}
\end{tabular}
\end{table}

\subsection{Buoyancy System}
The Buoyancy System consists of foam-filled screw shells and several Torus-shaped inflatables (referred to as bladders) placed over the U-joints. 

To produce a controllable buoyancy system, the following steps were taken: (1) Initial buoyancy calculations and tests to find the density and buoyant force for each segment, (2) A calculation to determine the passive buoyancy required for the segment buoyancy to be -5\%, meaning the segment will slowly sink, (3) A calculation to determine what volume of active buoyancy controlled through air filled bladders was needed to get to 5\% buoyant, for resurfacing. (4) The materials and pattern making to ensure a 2D textile would inflate to a desired 3D volume. (5) Subsequent tubing and upstream pressure were designed to work at the surface level and depth, minimizing the chance for bladder rupture.


\subsubsection{Buoyancy Calculations}
The system ideally has a buoyancy shift of ~5\% above and below neutrally buoyant to allow for control of floating, sinking, and swimming. To begin, a calculation was done to ascertain the buoyancy of the segments. 

\begin{equation}
\label{eq: buoy}
    F_b = \rho_w g V_{disp}
\end{equation}

where $F_b$ is the buoyancy force, $\rho_w$ the density of water, $g$ the gravitational constant, and $V_{disp}$ is the volume of water displaced by the segment. The highest-density segments are in the middle, with a net force of -19.4 N, making them 72\% neutrally buoyant. The other segments (front and tail) without the second half U-joint are both buoyant. In Table \ref{tab:MassandVol} in the mass column, there is an additional number provided, which is the weight added to the segment to make it dense enough to sink.


\subsubsection{Flotation Screw Shells}
To add passive buoyancy, the shells were given a larger volume and packed with a 2-part polyurethane Marine Foam. The addition of these shells to the middle segments put them at -10.4 N or 84\% neutrally buoyant. The size increase of the shells was limited due to the increase in diameter and mass, which negatively impacts the torque of the screw motor. The remaining volume would be handled through the bladders. The shell and the relevant buoyancy information can be found in Table \ref{tab:MassandVol}.

\subsubsection{Bladder Volume}
The buoyancy force contributed by each swim bladder is equal to the weight of water displaced, as shown in equation \ref{eq: buoy}. To calculate the buoyancy of the bladders, which are tarus-shaped, the Volume equation below was used:
\begin{equation}
    V_b = 2\pi^2r^2R
\end{equation}
where $V_b$ is the total volume for one bladder, $r$ is the minor diameter of the swim bladder, and $R$ is the major diameter of the swim bladder.

According to the buoyancy calculations for the middle segments, a buoyancy force of \raisebox{0.5ex}{\texttildelow}10 N is required for neutral buoyancy and \raisebox{0.5ex}{\texttildelow}13.1 N for 5\% buoyant. This indicates a target volume required of each bladder to be 0.00143 $m^3$ with a buoyant force of \raisebox{0.5ex}{\texttildelow}13.8 N, adding a 5\% buffer to the required buoyant force set point.

\subsubsection{Bladder Manufacturing} 

Nylon Taffeta Fabric was chosen for fabrication, as it is waterproof and airtight, ideal for creating the inflatable bladders. One side of the fabric is coated with a thermoplastic, enabling heat-bonding, while the other side is scratch-resistant. This ensures that the bladders are resistant to punctures when ARCSnake traverses a rocky environment. The swim bladder is constructed as a flat textile ring that becomes three-dimensional when filled with air. To design a bladder, the change in geometry needs to be accounted for,
\begin{align}
    D_{out} = D_{maj} + \frac{\pi}{2}D_{min} \:\:\:\:\:
    D_{in} = D_{maj} - \frac{\pi}{2}D_{min}
\end{align}

where $D_{maj}$ is the radius of the torus, $D_{min}$ is the diameter of the tube that makes up the torus, and $D_{out}$ and $D_{in}$ returns the required diameters of the textile ring. This defines a flat pattern that is larger in inner and outer diameter, becoming accurate when the volume is pressurized and forms a 3D geometry.

\subsubsection{Pressure System Calculations}
\label{sec: pressure calc}

Using separate pneumatic lines, the bladders at the front and rear of ARCSnake can be inflated or deflated independently, enabling underwater tilt control as shown in Fig \ref{fig: big figure} (D). The lower limit for the pressure reservoir supplying the buoyancy system with air is determined by setting the maximum fill time to 60s and calculating the pressure loss across the system, including head loss and pressure loss in the tubing. As the system is split into two branches (front and back), a pressure loss across each is calculated:

\begin{equation}
    P_{loss} = \underbrace{\frac{\rho_a f_DL}{D} \frac{\overline{v}^2}{2g}}_{Tubing\:Prssure\:Loss}
    + \underbrace{\sum K_i\frac{\overline{v}^2}{2g}}_{Head\:Loss}
\end{equation}

Where $P_{loss}$ is the total pressure loss in the system, all instances of $K_i$ are coefficients of minor losses [entrance loss: 0.8, split or branch: 2.0, valves: .05, exit loss: 1.0], $\rho_a$ (1.814$\frac{kg}{m^2}$) is the density of air at 25\degree C, $L$ ([3.58m, 4.11m]) is the length of the tubing for the back and front bladders, $D$ (vairable) is the inner diameter of the tube, $g$ (9.81$\frac{m}{s^2}$) is gravitational acceleration, $f_D$ is the friction factor of the tube and $\overline{v}$ is the mean velocity of each branch. The average velocity is found using the mass transport equation, and our friction factor, $f_D = \frac{64}{Re}$, is calculated based on the Reynolds number:

\begin{equation}
    \overline{v} = \frac{V_{b}}{t_{max}A} \;,\; Re = \frac{\rho_a v D}{\mu}
\end{equation}

Where $V_{b}$ (0.00141$\frac{m}{s^2}$) is the volume of the bladders, $t_{max}$ (60 s) is the max fill time, $A$ ($\pi D$) is the cross sectional area of the tubing, $Re$ is the Reynolds number, and $\mu$ (1.85$e^{-5}$) is the kinematic viscosity of air at 25\degree C. $Re$ is in a range to assume laminar flow, hence, a laminar flow profile.

All controllable variables are assumed to be fixed, except upstream pressure and $D$. The bladders used have an upper pressure limit of $\sim$7 psi, at which the bonded seam can rupture; therefore, 2.15 psi serves as a safe operating pressure.  Each bladder has a 6 mm Push-to-Connect Tube Fitting for Air used to connect it, ensuring the tubing fits under the screw shell. To maximize wall thickness, an inner diameter of 2 mm tubing was chosen (6 mm OD, 2 mm ID UV-Resistant Firm Plastic Tubing for Air and Water) and the "Y" and straight 6 mm push-to-connect fittings are from HOODUCTRIC. 

With these assumptions, a minimum upstream pressure of 2.9 psig is required, as noted in Fig. \ref{fig: big figure} (D), where 3-5 psi are used to control volume using pressure up to a 2 m depth. Upstream pressure is regulated using 2 Bellofram regulators set to 6 psi for the positive pressure line and 3-5 psi for the bladders, and a Kobalt Quiet Tech air compressor set to 15 psi tool pressure. 

\begin{figure*}[t]
    \centering
    \includegraphics[trim={0.5cm 4.5cm 1cm 10.5cm}, clip, width=.98\linewidth]{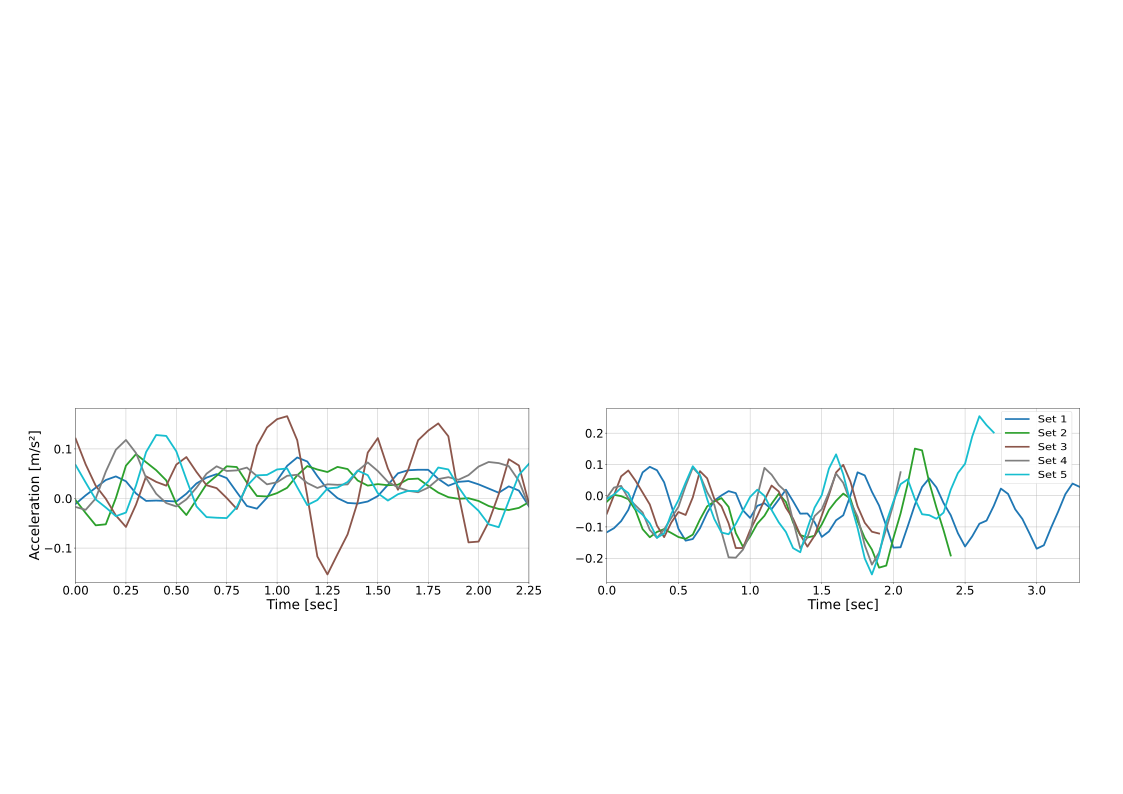}
    \caption{The ascent and descent accelerations were calculated through video post operations by selecting 5 key points across the 4 segments and the head of ARCSnake and using the time between frames to gather positional data at time steps differentiated twice to gather acceleration data. The acceleration, shown in the graphs, exhibits nearly constant acceleration in both directions, with average accelerations of 0.02717 $\pm$0.04522 m/s$^2$ and -0.04965 $\pm$ 0.04332 m/s$^2$ respectively.}
    \label{fig:buoyancy_bladder_system}
\end{figure*}


\begin{figure}
    \vspace{2mm}
    \centering
    \includegraphics[trim={4cm 0 4cm 0}, clip, width=0.48\textwidth]{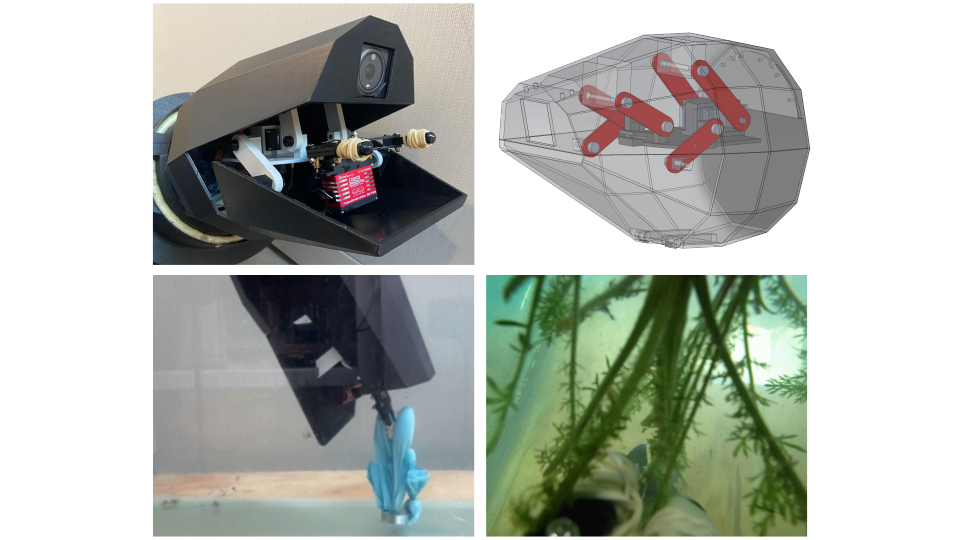}
    \caption{The top left image displays ARCSnake's head in open configuration. The top right images shows the 4 bar linkage when the mouth is closed. The bottom left image shows the head grasping trash underwater. The bottom right image shows a first-person view of the claw collecting a botanical specimen.}
    \label{fig:snake_head}
\end{figure}

\subsection{Electronics and Communication}


\subsubsection{Power Electronics}
Includes power tether, penetrators, and PCB/Converters.
The ARCSnake power system comprises a 48V power rail with remote sense and power electronics individually housed in each segment. The 48V power rail is implemented with a BlueRobotics High-Temperature PUR Subsea Cable, P/N BR-101093 cable, and BlueRobotics WetLink penetrators for watertight construction. An SJEOOW conductor, rated for operating in outdoor and water conditions, is used for remote sensing. \\ 
The power electronics individually housed in each segment are managed by a custom-printed circuit board (PCB). The PCB operates a DC-DC converter, VICOR V48B24C250BL, to distribute 24V power from the 48V rail. The U-joint motors associated with each segment are powered directly from the 48V rail. The PCB distributes 24V to the screw shell motor, microcontroller, servo motors, and communication elements.


\subsubsection{Communication}
The snake communication is based on CANBus, where each component (e.g., motors) has a unique address that communicates to the host PC.
A Fathom ROV Tether from Blue Robotics was used to connect the CAN line between the PC and the snake system.
Between the segments, a waterproof shielded ~20 gauge wire was run that enters then exits each segment using the wetlink penetrator from Blue Robotics.
In each segment, the CAN line splits to connect with all motors, two or three, depending on whether it is a middle or end segment, and the CANBed - Arduino CAN Bus Dev Kit.
The Arduino board is then connected to the internal IMU through I2C, which can be used to gather data on relative motions or angles between the segments or their orientation relative to gravity.

\section{Experiments}

\subsection{Screw Drive Validation}
The maximum torque produced by the motor during closed-loop speed control increases with the commanded speed. The peak tangential force measured ranged from 40.0 N to 75.9 N at commanded speeds of 10 rad/s and 50 rad/s, respectively. This corresponds to a resultant torque ranging from 3.60 to 6.83 Nm. Given that the torque output of the motor with no load is about 1.5 Nm maximum and the gear ratio of the screw drive train is 7:1, the ideal resultant torque output at the screw shell should be 10.5 Nm. Thus, the efficiency of the screw drive train is calculated to be about 65.7\% at a commanded speed of 50 rad/s. 

\subsection{Buoyancy System}
\subsubsection{Volume Measurements}
Given the complexity of the inner chamber and U-joint attachments, a water displacement test was used to calculate the volume of the segments. A known volume of water was placed in a rectangular container. The water height was measured before and after a middle segment was fully submerged. The segment volume is listed in Table \ref{tab:MassandVol}.

\subsubsection{Buoyancy Control Tests}
To validate our calculations in sec \ref{sec: pressure calc}, an experiment was conducted trying to match the expected fill time at atm. The bladders started fully deflated. Both valves connected to 2.9psi were opened simultaneously. In 68s, the rear bladders inflated and the front in 70s. This is a 13.3\% error from the desired inflation time. 

\subsubsection{Sinking and Rising} 
To test the buoyancy system, the snake was first placed on the water surface with the bladders deflated. The snake was dropped while being filmed from a side-view window. Once the snake reached the bottom, the pressure line to all bladders was opened, and the snake was again filmed rising for post-video processing. 

The results of the experiment, as shown in Fig. \ref{fig:buoyancy_bladder_system}, are as follows: When the bladders are deflated, ARCSnake fell with an average acceleration of 0.045 $\pm$ 0.04332 m/s$^2$. In field testing, ARCSnake took approximately 10 seconds to sink to the bottom of a 1.5-meter water tank. Once at the bottom, the bladders were reinflated with an upstream pressure of 6 psi, and ARCSnake rose with an acceleration of 0.02717 $\pm$0.04522 m/s$^2$.

\subsection{U-Joint Validation}



\begin{figure} [t!]
    \centering
    \includegraphics[width=\linewidth, trim={10 32 0 0}, clip]{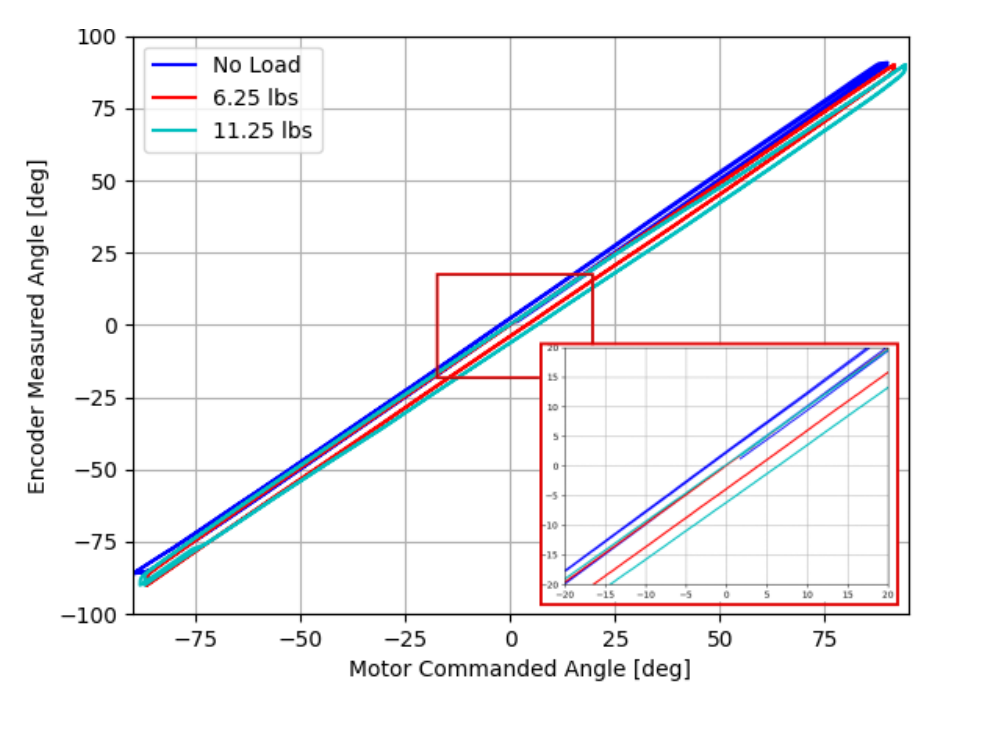}
    \caption{The graph above depicts the input command vs measures the angle of the motor with varying loads applied. With no load (blue), the hysteresis loop has a width of \raisebox{0.5ex}{\texttildelow}2 degrees, at 6.25 lbs (red), the width increases to \raisebox{0.5ex}{\texttildelow}4 degrees, and at 11.25 lbs of load (cyan), the width reaches \raisebox{0.5ex}{\texttildelow}6 degrees.}
    \label{fig:u-joint data figure}
\end{figure}

The U-joint system is validated through a repeatability test, in which the joint follows several periods of a sine-wave trajectory. Varying weights were placed on the end of a half u-joint, and the joint was prescribed a repeated pattern to and from its joint limits to measure hysteresis. The results are shown in Fig. \ref{fig:u-joint data figure}. With no load (blue), the hysteresis loop has a width of \raisebox{0.5ex}{\texttildelow}2$^\circ$, at 6.25 lbs (red), the width increases to \raisebox{0.5ex}{\texttildelow}4$^\circ$, and at 11.25 lbs of load (cyan), the width reaches \raisebox{0.5ex}{\texttildelow}6$^\circ$.

\section{Discussion and Conclusion}

This work presents ARCSnake V2, a screw-propelled, water-sealed adaptation of ARCSnake V1 with amphibious capabilities for aquatic environments. By integrating screw-based propulsion with hyper-redundant joint actuation and a modular, waterproofed architecture, ARCSnake V2 demonstrates effective multi-domain mobility. Experimental results validate its mechanical efficiency, force regulation capabilities, and precise maneuverability in submerged conditions. ARCSnake was deployed in a water tank as seen in Fig. \ref{fig:cover_figure}, showcasing its effectiveness in water. Another field test was completed with the head, where the claw grasped aquatic specimens and trash from the bottom of the tank, seen in Fig. \ref{fig:snake_head}.

The platform’s novel combination of serially linked screw drive segments, cable-driven universal joints, and adaptive buoyancy control enables a diverse set of locomotion modes, supporting its use in unstructured and hazardous environments. These capabilities position ARCSnake V2 as a versatile platform for tasks ranging from sample collection and underwater trash removal to pipe inspection and search-and-rescue missions. For instance, the robot can be deployed to grasp objects on a riverbed with its claw and then use its buoyancy system for retrieval on the surface. Similarly, it can wrap around pipes for inspection while minimizing disruption to surrounding wildlife, and its screw-based propulsion is inherently non-damaging to most surfaces while being less prone to debris entanglement than traditional propeller-driven systems.

Future work will enhance autonomous control strategies through sensor fusion and expand the platform’s scalability for more complex mission profiles. These advancements aim to further improve the system’s adaptability and effectiveness in real-world exploration and intervention tasks.

\section{Acknowledgements}
We want to thank Mingwei Yeoh, Colin Brady, Kylie Chan, Andrew Nakoud, Anne-Marie Shui, Yaohui Chen, Zoe Samuels, Diego Williams, Tarun Murugesan, Derek Chen, Dimitri Schreibier, Hoi Man (Kevin) Lam, and Casey Price for their contributions.

\bibliographystyle{ieeetr}
\bibliography{refs}

\newpage

 





\end{document}